\documentclass[sigconf]{acmart}
\AtBeginDocument{%
  }

\renewcommand\footnotetextcopyrightpermission[1]{}

\setcopyright{acmlicensed}
\copyrightyear{2018}
\acmYear{2018}
\acmDOI{XXXXXXX.XXXXXXX}
\acmConference[]{Preprint}{January, 2026}{}
\usepackage{graphicx}
\usepackage{booktabs}
\usepackage{enumitem}
\usepackage{multirow}
\usepackage[table, dvipsnames, svgnames, x11names]{xcolor}
\usepackage{pifont} % 用于使用 ding 字符集
\usepackage{tikz}
\usepackage{graphicx} % 引入图片宏包
\usepackage{subcaption} % 引入子图宏包
\usepackage{makecell}
\usepackage{amsthm}
\usepackage{marvosym}

\definecolor{darkgreen}{RGB}{0,100,0} % 自定义深绿色
\definecolor{lightgreen}{RGB}{180, 220, 180} % 自定义浅绿色
\usepackage[figuresright]{rotating}
\usepackage[most]{tcolorbox}
\newtcbox{\mybox}[1][red]
  {on line, arc = 0pt, outer arc = 0pt,
    colback = #1!10!white, colframe = #1!50!black,
    boxsep = 0pt, left = 1pt, right = 1pt, top = 2pt, bottom = 2pt,
    boxrule = 0pt, bottomrule = 1pt, toprule = 1pt}
\definecolor{BoxBackground}{RGB}{243, 242, 236} % 浅灰色背景
\definecolor{BoxFrame}{RGB}{255, 255, 255} % 黑色边框
\definecolor{TitleBackground}{RGB}{230,39,39} % 标题背景颜色
\definecolor{TitleText}{RGB}{255, 255, 255} % 标题文字颜色
\tcbset{
  academicbox/.style={
  width=\textwidth, % 设置盒子宽度为两栏总宽度
    halign=left, % 水平居中对齐
    bottom=0.5pt,
    colback=BoxBackground,
    colframe=BoxFrame,
    colbacktitle=TitleBackground,
    coltitle=TitleText,
    enhanced,
    attach boxed title to top left={yshift=-0.1in,xshift=0.1in},
    boxed title style={boxrule=0pt,colframe=white},
    title={#1},
  }
}
\newtcolorbox{AcademicBox}[1][]{academicbox=#1}

\definecolor{SoftBlue}{RGB}{135, 206, 250}  % 浅蓝色
\definecolor{SoftOrange}{RGB}{255, 224, 178} % 浅橙色
\definecolor{SoftGreen}{RGB}{144, 238, 144}  % 浅绿色
\definecolor{CorrectGreen}{RGB}{76, 175, 80} % 淡绿色，适用于表示正确
\definecolor{ErrorRed}{RGB}{211, 47, 47} % 深红色，适用于表示错误

\newcommand{\ep}{$\texttt{PROPHET}$}
\newcommand{\ace}{\texttt{CIL}}
\usepackage{listings}

\definecolor{colorBACKGROUND}{HTML}{F3F2EC} % 浅米白
\definecolor{colorKEYWORD}{HTML}{1E93AB}    % 深青色
\definecolor{colorSTRING}{HTML}{E62727}     % 鲜红色
\definecolor{colorFRAME}{HTML}{DCDCDC}      % 浅灰色

\lstdefinestyle{CustomStyle}{
    language=Python,
    backgroundcolor=\color{colorBACKGROUND}, 
    keywordstyle=\color{colorKEYWORD}\bfseries,
    stringstyle=\color{colorSTRING},
    commentstyle=\color{gray}, % 保持注释为标准灰色，以避免与数据混淆
    numberstyle=\tiny\color{colorFRAME},
    basicstyle=\small\ttfamily,
    breaklines=true,
    showstringspaces=false,
    tabsize=4,
    frame=single, 
    framerule=0.5pt,
    rulesepcolor=\color{colorFRAME},
    numbers=left,
    stepnumber=1,
    numbersep=5pt,
    captionpos=b,
}

\usepackage{amsmath,amsfonts,bm}

\def\eqref#1{equation~\ref{#1}}
\def\1{\bm{1}}

\DeclareMathAlphabet{\mathsfit}{\encodingdefault}{\sfdefault}{m}{sl}
\SetMathAlphabet{\mathsfit}{bold}{\encodingdefault}{\sfdefault}{bx}{n}

\def\sR{{\mathbb{R}}}

\def\sX{{\mathbb{X}}}

\makeatletter
\renewcommand*{\@fnsymbol}[1]{\ensuremath{\ifcase#1\or \dagger\or \ast\or \ddagger\or
   \mathsection\or \mathparagraph\or \|\or **\or \dagger\dagger
   \or \ddagger\ddagger \else\@ctrerr\fi}}
\makeatother

\begin{document}

\title{\ep: An Inferable Future Forecasting Benchmark with Causal Intervened Likelihood Estimation}

\author{Zhengwei Tao}
\email{tttzw@pku.edu.cn}
\affiliation{%
  \institution{Peking University}
\country{China}
}

\author{Pu Wu}
% \authornote{Corresponding authors.}
\email{puwu1997@126.com}
\affiliation{%
  \institution{Peking University}
\country{China}
}

% --- 第3位作者：Zhi Jin ---
\author{Zhi Jin}
% 定义脚注内容：这里写复数形式，它会分配一个符号（如果这是全文第一个note，默认是*；如果前面有同等贡献，这里会自动变为†）
\authornote{Corresponding authors.} 
\email{zhijin@whu.edu.cn}
\affiliation{%
  \institution{Wuhan University}
  \country{China}
}

% --- 第4位作者：Xiaoying Bai ---
\author{Xiaoying Bai}
% 关键点：使用 authornotemark[1]
% 这表示“使用与第1个 authornote 相同的符号”。
% 这样两人名字右上角符号完全一致，且页脚只会显示一次 "Corresponding authors."
\authornotemark[1] 
\email{baixy@aibd.ac.cn}
\affiliation{%
  \institution{AIBD}
  \country{China}
}

\author{Haiyan Zhao}
\email{zhhy@sei.pku.edu.cn}
\affiliation{%
  \institution{Peking University}
\country{China}
}

\author{Chengfeng Dou}
\email{chengfengdou@pku.edu.cn}
\affiliation{%
  \institution{Peking University}
\country{China}
}

\author{Xiancai Chen}
\email{xiancaich@stu.pku.edu.cn}
\affiliation{%
  \institution{Peking University}
\country{China}
}

\author{Jia Li}
\email{jia.li@whu.edu.cn}
\affiliation{%
  \institution{Wuhan University}
\country{China}
}

\author{Linyu Li}
\email{linyuli@stu.pku.edu.cn}
\affiliation{%
  \institution{Peking University}
\country{China}
}

\author{Chongyang Tao}
\email{chongyang@buaa.edu.cn}
\affiliation{%
  \institution{Beihang University}
\country{China}
}

\author{Wentao Zhang}
\email{wentao.zhang@pku.edu.cn}
\affiliation{%
  \institution{Peking University}
\country{China}
}

% \affiliation{%

% \affiliation{%

% \affiliation{%

% \affiliation{%

% \affiliation{%

% \affiliation{%

% \affiliation{%

\renewcommand{\shortauthors}{Zhengwei Tao et al.}

\begin{abstract}
Predicting future events based on news on the Web stands as one of the ultimate aspirations of artificial intelligence. Recent advances in large language model (LLM)-based systems have shown remarkable potential in forecasting future events, thereby garnering significant interest in the research community. 
Currently, several benchmarks have been established to evaluate the forecasting capabilities
by formalizing the event prediction as a retrieval-augmented generation (RAG)-and-reasoning task. In these benchmarks, each prediction question is answered with relevant retrieved news articles downloaded from the Web.
However, because there is no consideration of whether the questions can be supported by valid or sufficient supporting rationales, some of the questions in these benchmarks may be inherently noninferable.
To address this issue, we introduce a new benchmark, \ep, which comprises inferable forecasting questions paired with relevant news for retrieval. 
To ensure the inferability of the benchmark, we propose Causal Intervened Likelihood (\ace), a statistical measure that assesses inferability through causal inference. 
In constructing this benchmark, we first collected recent trend forecasting questions, 
and then filtered the data using \ace~resulting in an inferable benchmark for future forecasting. 
Through extensive experiments, we first demonstrate the validity of \ace~ and in-depth investigations into future forecasting with the aid of \ace. Subsequently, we evaluate several representative prediction methods on \ep. The overall results draws valuable insights for task of future directions.
Data is public on \url{https://github.com/TZWwww/PROPHET}.
\end{abstract}

\settopmatter{printacmref=false} % 关闭 ACM reference format
\maketitle

\vspace{-2mm}

\section{Introduction}

The quest to forecast future events based on information on the Web has long been a central pursuit in the field of artificial intelligence (AI). The ability to foresee outcomes and trends holds the promise of revolutionizing numerous sectors covering finance~\cite{li-etal-2024-alphafin}, climate science~\cite{wang2024exploring}, and social policy~\cite{rotaru2022event}. Recent years have witnessed a surge in interest and progress, particularly with the advent of large language model (LLM)-based systems. These systems, leveraging the power of deep learning and vast amounts of data, have demonstrated an unprecedented capacity for forecasting, capturing the imagination and focus of the research community~\cite{halawi2024approaching, hsieh2024reasoning, pratt2024can}.

\begin{figure}
\setlength{\belowcaptionskip}{-5mm}
    \centering
    \includegraphics[width=1\columnwidth]{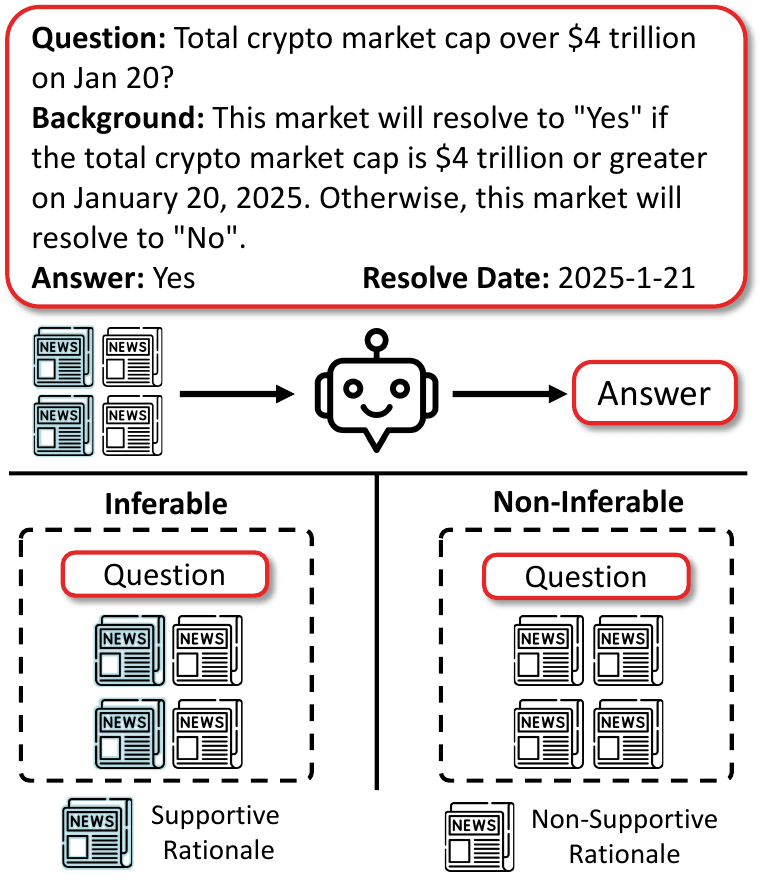}
    \caption{The upper Figure demonstrates the task of future forecasting. The lower half shows both inferable and non-inferable scenarios.}
    \label{fig:intro}
\end{figure}

To evaluate the abilities of these LLM-based future forecasting systems, pilot works construct several benchmarks based on real-world forecasting questions~\cite{halawi2024approaching, guan2024openep, karger2024forecastbench,zeng2025futurex}.
These benchmarks have successfully framed future forecasting as a retrieval-augmented generation (RAG)-and-reasoning task.
Within this framework, systems should first search the Web or databases for news articles related to the prediction question in the benchmarks to gain knowledge base, then reason based on the retrieved knowledge base.
Nevertheless, in order to truly evaluate the abilities of the LLM-based future forecasting, the prediction questions in the benchmarks need to be inferable,
meaning that the supporting knowledge base must contain sufficient information to substantiate the answers. 
In traditional RAG tasks, the answer can definitely be found within the knowledge base. 
However, future forecasting tasks do not inherently satisfy this characteristic compared to traditional knowledge-intensive benchmarks such as HotpotQA~\cite{yang2018hotpotqa} and 2WikiMultiHopQA~\cite{ho2020constructing}.
That is, future forecasting needs to be inferred by rationales, i.e. facts and reasoning clues, but the knowledge base may only provide partially supportive rationales for the prediction questions~\cite{zhao2024retrieval}.
Collecting real-world prediction questions as the benchmark without nuanced validation, the knowledge base may not be able to provide sufficient supportive facts which makes some of the prediction questions non-inferable~\cite{birur2024vera}. 

To overcome this challenge, we introduce an inferable future forecasting benchmark, \ep, designed to provide a more accurate evaluation. To ensure reproducibility, \ep~is an RAG task where each real-world prediction question pairs with relevant downloaded news articles for retrieval from the Web.
We are next motivated to select prediction questions that are inferable, based on their related articles.
The most challenging part is to estimate the inferability of each question since we cannot observe the completed real-world event evolution process. Even if we can, it is difficult to determine as well, due to the lack of expert knowledge of a wide spectrum of domains.
A key innovation in our approach is the introduction of Causal Intervened Likelihood (\ace), a statistical measure that assesses the inferability of prediction questions through causal inference. \ace~is calculated via principles of causal inference where we measure the supporting degree of each article for the answer to the question. We regard each article as an event and compute the effect of intervening in the event from happening to not happening.
\ace~provides a robust estimate of whether a question can be answered. 
We then filter the prediction questions using \ace~to ensure the inferability of the benchmark, providing a fair and accurate evaluation of the systems' forecasting ability.
Assisted by \ace, \ep~performs as a well-formulated RAG-and-reasoning task with hidden rationale~\cite{zhao2024retrieval}.

To validate the effectiveness of \ace~, we conducted a series of extensive experiments. These experiments were designed to rigorously test how this estimation can represent the inferability of prediction questions. The results of the experiments were highly encouraging, demonstrating a strong correlation between \ace~scores and the actual performance of the systems in terms of both retrieval and prediction accuracy.
Further, \ace~enables us to conduct in-depth investigations into future forecasting, drawing out innate properties of this complicated task.
Finally, we evaluated several state-of-the-art prediction systems on the \ep~benchmark. This evaluation provided effective measurements of the strengths and weaknesses of each system, highlighting areas for improvement and potential directions for future research. 
We will also regularly update the dataset to ensure its timeliness and to minimize the risk of data leakage due to model evolution.
To summarize our contribution:
\begin{itemize}[left=0.2cm]\setlength{\itemsep}{0pt}\setlength{\parskip}{0pt}\setlength{\topsep}{-5pt}

\item[$\bullet$] We are the first to introduce \ace~for inferability estimation of the future forecasting. We provide a feasible method for calculating this metric, which we also testify to its validity..

\item[$\bullet$] Assisted by \ace, we establish an automatic pipeline to construct the future forecasting benchmark \ep~where the real-world prediction questions are insufficiently inferable based on their related news articles. 
% We will keep updating it with the pipeline once new questions arise.

\item[$\bullet$] We evaluate several baselines for future forecasting. The results show the pros and cons of these systems and present great potential and development directions for this task. 

\end{itemize}

% \vspace{-3mm}
\section{Preliminaries}

% \vspace{-3mm}
\subsection{Future Forecasting}
Future forecasting stands for predicting whether a certain event will happen in the future based on news information from the Web. We now formalize the task as a binary question-answering task. Given a prediction question $Q$ which can be ``Will Tim Walz win the VP debate against J.D. Vance?'' or ``Will Bitcoin rise to \$100,000 by December 2024?''. 
There would be background information $B$ that describes the context of $Q$.
A large set of news articles $\sX$ serves as a knowledge base to retrieve. 
The forecasting system must answer the question as formalized:
\setlength\abovedisplayskip{1.5pt}
\setlength\belowdisplayskip{1.5pt}
\begin{equation} 
\begin{aligned}
        Y = \mathrm{Reason}(Q, B, \sX),
\end{aligned}
\end{equation}
where $Y\in[0, 1]$ is the predicted probability of how likely the event in $Q$ would occur. 
A ground truth answer
$\hat{Y}\in\{0, 1\}$ paired with a resolved date $D$ represents whether the event in $Q$ finally occurs and the date the question resolves. 
As the same in previous works~\cite{halawi2024approaching, karger2024forecastbench}, we use Brier Score~\cite{brier1950verification} as the metric for evaluation:
\begin{equation} 
\begin{aligned}
        \texttt{Brier Score} = \frac{1}{N}\sum_{n}^{N} (Y_n - \hat{Y}_n)^{2},
\end{aligned}
\end{equation}
$N$ is the number of questions in the dataset. In this work, we report the Brier Score expanded by 100 times and keep 2 decimal places.

We formalize future forecasting as an RAG task. As an RAG, it features distinctly compared with traditional dataset such as HotpotQA~\cite{yang2018hotpotqa} and 2WikiMultiHopQA~\cite{ho2020constructing}.
The knowledge base $\sX$ stores the rationales and clues for answering $Q$~\cite{zhao2024retrieval}.
Future forecasting mainly detects two core entangled abilities of the systems: retrieval and reasoning.

Current future forecasting benchmarks are constructed by harvesting real-world prediction questions and paired with news articles before the resolved date $D$~\cite{halawi2024approaching, guan2024openep, karger2024forecastbench} without nuanced validation of the inferability of the questions.
It is possible that there is a lack of sufficient supportive information in $\sX$ for the question. 
Methods need to be established to ensure that the prediction questions in the benchmarks are sufficiently inferable.

% \vspace{-4.2mm}

\subsection{Causal Inference} \label{sec: causal_inference}
Causal inference is a vital statistical method to determine causal relationships between variables~\cite{{pearl2010introduction}}. In real-world scenarios, a mere correlation between two variables may be due to chance or hidden factors. Causal inference aims to establish direct causality.
For example, the increase in ice cream sales and drowning incidents is not a causal link, although both are affected by hot weather.
Causal inference uses concepts such as structural causal models, interventions, and counterfactual inferences. These are applied in medicine, economics, and social sciences.

\noindent\textbf{Structural causal model (SCM)} 
It is a framework designed to represent and analyze causal relationships between variables using a combination of causal graphs and structural equations. At its core, SCM relies on a directed graph where nodes represent variables $X$, and edges denote direct causal influences, forming a network that captures dependencies and pathways of causation. 
Each variable in the model is determined by its direct causes (parent nodes). SCM enables the identification of causal effects, and the exploration of intervention questions (e.g., "What would happen if we intervened on X?"). This has been widely applied in fields like epidemiology, economics, and machine learning to disentangle complex causal mechanisms and validate hypotheses~\cite{stolfo2023causal}.

\noindent\textbf{Interventional distribution}
An SCM allows the study of interventions. An atomic intervention $\text{{do}}(X_i = x)$ fixes $X_i$ with a fixed value $x$. 
For example, in a medical trial, the dose of a new drug is set at a specific value for a group.
In the view of the structural causal model, interventions can be understand as changing of the original structure and variable distributions. After $\text{{do}}(X_i = x)$, the resulting distribution is $P(\cdot|\text{{do}}(X_i = x)) \doteq P_{{m}}(\cdot|X_i = x)$, which shows how other variables respond. 

A critical distinction in causal inference lies between interventional and observational probabilities. The observational probability, denoted as the conditional probability $P(Y|X=x)$, represents the likelihood of event $Y$ given that we have passively observed variable $X$ to be $x$. This probability captures mere statistical association or correlation. In contrast, the interventional probability, denoted as $P(Y|\text{do}(X=x))$, represents the likelihood of $Y$ if we were to actively intervene and set the value of $X$ to $x$. This distribution describes the causal effect of $X$ on $Y$ by simulating a controlled experiment and removing spurious correlations. While interventional probabilities are essential for predicting the effects of actions, they cannot be computed directly from passively collected data. Therefore, a primary challenge in causal inference is to determine whether and how these interventional probabilities can be calculated from available observational probabilities\cite{pearl2010introduction}.

% \vspace{-3mm}
\section{\ep~Benchmark}
% \vspace{-3mm}
In this section, we introduce \ep~ which is a future forecasting benchmark with inferability estimation and selection. We first describe the data collection process in Section~\ref{sec: data_collection}. Then we introduce the Causal Intervened Likelihood (\ace) metric in Section~\ref{sec: CIL}. We describe the benchmark construction in Section~\ref{sec: construction} and ~\ref{sec:composition}. We report the statistics of \ep~in Section~\ref{sec:stats}.

\subsection{Data Collection}
\label{sec: data_collection}

Our objective is to gather a dataset that encompasses recent and prominent prediction questions. To achieve this, we have sourced questions from well-known platforms: Polymarket\footnote{https://polymarket.com/}. The domains covered by the questions on these platforms are highly diverse, ranging from scientific breakthroughs to social and economic trends. This diversity ensures that the benchmark is representative of a wide spectrum of forecasting tasks. Moreover, the questions are trending and among the most attention-attracting ones on the platform. This indicates that they are not only relevant in the current context but also likely to be of interest to the broader forecasting community. As such, the data collected from these sources provides a robust foundation for evaluating and developing practical forecasting models.  

To avoid model leakage, we carefully selected questions. We aim to use the latest forecast questions.
We include all questions on Polymarket resolved between 2025,1,1 and 2025,1,31.
We filter out meaningless questions, such as personal inquiries or those with little community interest, to focus on realistic forecasting scenarios.
After collecting questions, we collected relevant news articles as much as possible. Using LLM, we generated three types of news search queries per question: entities in the question, resolving steps, and similar historical events using prompts in the Appendix~\ref{app: prompt} (a-c).
Then we searched on the MediaCloud open-source platform\footnote{{https://www.mediacloud.org}} with these queries. MediaCloud's vast news repository helped us gather comprehensive news URLs.
After gathering the news URLs, we use Newspaper3k\footnote{https://newspaper.readthedocs.io/} API to download the news articles. We discard news that is later than the resolved date of the question.

However, many downloaded articles were irrelevant. There are also news articles tagged with wrong publish date, which could incur data leakage.
To address these, we use LLM to rate the relevant score of each article to the question. We set the highest score, indicating this article directly answering of the question, while the lowest score showing total irrelevance. We remove both articles with the highest and lowest scores. The relevance rating prompt is in the Appendix~\ref{app: prompt} (d).

\begin{figure}
\setlength{\belowcaptionskip}{-3mm}
\setlength{\abovecaptionskip}{-2mm}
    \centering
    \includegraphics[width=1\columnwidth]{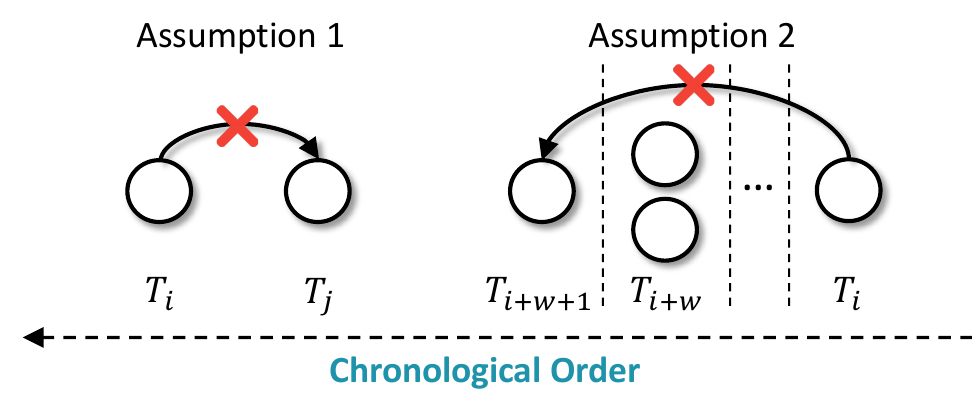}
    \caption{Illustration of assumptions. Nodes are variables that are in chronological order with time $T$.}
    \label{fig:assumption}
\end{figure}

\begin{table*}
\centering
\setlength{\abovecaptionskip}{-3mm}
  \setlength\tabcolsep{5pt}
\begin{tabular}{cccccccc}
\toprule
\# Question & \# News & Avg Article Word & Avg \ace & Avg Pos \ace & Avg Neg \ace & Earlies News Date & Latest News Date \\
\midrule
612  & 559.51  & 1249.88 & -0.013 & 0.2041 & -0.2503 & 2023,1,1 & 2025,1,1 \\
\bottomrule
\end{tabular}
\caption{Statistics of L1 part of \ep. }
\label{tab:stats}
\end{table*}

\subsection{Causal Intervened Likelihood}
\label{sec: CIL}
To measure the sufficiency of supportive rationales for each question and to construct an inferable benchmark, we introduce a statistical estimation named \textbf{Causal Intervened Likelihood (CIL)}, derived from causal inference. CIL estimates how strongly each news article supports the answer to a given question.

We use Bernoulli variables to model the occurrence of events. Let $Y \in \{0, 1\}$ indicate whether the event asked by the question occurs, and let $X_i \in \{0, 1\}$ indicate whether the event described in the $i$-th news article occurs. Each variable $X_i$ is associated with a happening date $T_i$. We use the notation $T_i \prec T_j$ to represent that the occurrence of the $i$-th news is before that of the $j$-th. Note that the date of $Y$ is later than any date of $X_i$.

Intuitively, if the occurrence of the $i$-th news article ($X_i=1$) is a necessary condition for the answer $Y=\hat{Y}$, then the intervention $\mathrm{do}(X_i=0)$ would significantly increase the probability of $Y \neq \hat{Y}$. With this intuition, we define the CIL of the $i$-th news article as:
\begin{equation}
\label{eq:ace}
\ace_i = P(Y=\hat{Y} | \mathrm{do}(X_i=1)) - P(Y=\hat{Y} | \mathrm{do}(X_i=0)),
\end{equation}
where $\mathrm{do}(\cdot)$ is the intervention operation from causal inference, representing that $X_i$ is forced to happen or not.

The \ace~estimation is composed by interventional probabilities.
As similar to the core idea in causal inference, we derive these interventional probabilities to observational probabilities. 
We model all $X_i$ and $Y$ as a Structural Causal Model (SCM), where variables are nodes and causal relationships are directed edges. However, determining the complete SCM is extremely challenging due to incomplete knowledge and the need for intensive expert analysis. Therefore, calculating CIL via methods that rely on a complete SCM is impractical.
Thus we introduce two key assumptions.

\noindent\paragraph{Assumption 1. Temporality} \textit{For any two news events, the one that occurs later cannot causally affect the one that occurred earlier:}
\begin{equation}
\forall i,j, \quad \text{if } T_i \prec T_j, \text{then} (X_j, X_i) \notin \text{edges~of~SCM}.
\end{equation}
This common-sense assumption aligns causal relationships with the flow of time and eliminates cycles in the SCM. Note that $Y$ is the chronologically last variable in this model.

% \vspace{0.5em}
\noindent\paragraph{Assumption 2. $w$-day Dependency Window} \textit{The direct causal influence between news events is time-limited. A news event $X_j$ can only have a direct causal effect on a news event $X_i$ if~$X_j$ occurs within a $w$-day window before $X_i$:}
\begin{equation}
\forall i,j, \quad \text{if } (T_i - T_j) > w \text{ days, then} (X_i, X_j) \notin \text{edges~of~SCM}.
\end{equation}
This assumption posits that the causal influence of distant past news events is mediated by more recent, intermediate news events.

With these assumptions, we can derive the CIL estimation. 
We first show the calculation for $P(Y=\hat{Y}|\mathrm{do}(X_i=1))$; the term $P(Y=\hat{Y}|\mathrm{do}(X_i=0))$ can be computed analogously.

\noindent\paragraph{Proposition.}  
\textit{The intervened probability $P(Y=\hat{Y} \mid \mathrm{do}(X_i=1))$ can be expressed purely in terms of observational probabilities:}  
\begin{equation}
\label{eq:proposition}
\begin{aligned}
P\big(Y=\hat{Y} \mid \mathrm{do}(X_i=1)\big) 
&= \sum_{\mathbf{x}_{\mathcal{N}_i}} 
    P\!\left(Y=\hat{Y} \mid X_i = 1,\ \mathbf{X}_{\mathcal{N}_i} = \mathbf{x}_{\mathcal{N}_i}\right) \\
&\quad\times P\!\left(\mathbf{X}_{\mathcal{N}_i} = \mathbf{x}_{\mathcal{N}_i}\right).
\end{aligned}
\end{equation}

\begin{itemize}[left=0.2cm]\setlength{\itemsep}{0pt}\setlength{\parskip}{0pt}\setlength{\topsep}{-5pt}
    \item $\mathcal{N}_i = \{\, j \mid 0 \le T_i - T_j \le w \,\}$ is the set of indices to news articles published within the $w$-day window prior to $X_i$.  
    \item $\mathbf{X}_{\mathcal{N}_i}$ denotes the vector of binary random variables $(X_j)_{j \in \mathcal{N}_i}$, where $X_j \in \{0,1\}$ indicates the occurrence of $j$th news event.  
    \item $\mathbf{x}_{\mathcal{N}_i}$ denotes one specific assignment of values to $\mathbf{X}_{\mathcal{N}_i}$, i.e., a binary vector $(x_j)_{j \in \mathcal{N}_i}$ with each $x_j \in \{0,1\}$.  
    \item The summation $\sum_{\mathbf{x}_{\mathcal{N}_i}}$ runs over all $2^{|\mathcal{N}_i|}$ possible binary value combinations of $\mathbf{x}_{\mathcal{N}_i}$, thereby enumerating every possible pattern of possibility of events in $\mathcal{N}_i$.
\end{itemize}

\begin{proof}
Let $\mathcal{N}_i = \{j \mid 0 \le T_i - T_j \le w\}$ be the set of indices for variables within the $w$-day window before $X_i$, and let $\mathcal{M}_i = \{j \mid T_i - T_j > w\}$ be the set for those outside the window. Using the law of total probability, we can expand the post-intervention probability by marginalizing over all variables that are not descendants of $X_i$. Due to Assumption 1 (Temporality), we only need to consider variables that precede $X_i$.
\begin{equation}
\label{eq1}
\begin{aligned}
& P(Y=\hat{Y}|\mathrm{do}(X_i=1)) \\
& = \sum_{\mathbf{x}_{\mathcal{N}_i}} \sum_{\mathbf{x}_{\mathcal{M}_i}} P(Y=\hat{Y}|X_i=1, \mathbf{X}_{\mathcal{N}_i}=\mathbf{x}_{\mathcal{N}_i}, \mathbf{X}_{\mathcal{M}_i}=\mathbf{x}_{\mathcal{M}_i}) \\
& \quad \times P(\mathbf{X}_{\mathcal{N}_i}=\mathbf{x}_{\mathcal{N}_i}, \mathbf{X}_{\mathcal{M}_i}=\mathbf{x}_{\mathcal{M}_i} | \mathrm{do}(X_i=1)).
\end{aligned}
\end{equation}
Based on the adjustment formula in causal inference~\cite{pearl2010introduction}, an intervention on $X_i$ only affects its descendants. Due to Assumption 1, the variables $\mathbf{X}_{\mathcal{N}_i}$ and $\mathbf{X}_{\mathcal{M}_i}$ all occur before $X_i$ and thus cannot be its descendants. Therefore, the intervention on $X_i$ does not affect their joint probability:
$P(\mathbf{X}_{\mathcal{N}_i}, \mathbf{X}_{\mathcal{M}_i} | \mathrm{do}(X_i=1)) = P(\mathbf{X}_{\mathcal{N}_i}, \mathbf{X}_{\mathcal{M}_i})$.

Furthermore, Assumption 2 states there are no direct causal paths from variables in $\mathbf{X}_{\mathcal{M}_i}$ to $Y$ that are not mediated by variables closer to $Y$ (including those in $\mathbf{X}_{\mathcal{N}_i}$ and $X_i$). This implies that given $X_i$ and its more immediate predecessors $\mathbf{X}_{\mathcal{N}_i}$, $Y$ becomes conditionally independent of the more distant predecessors $\mathbf{X}_{\mathcal{M}_i}$.
\begin{equation}
\label{eq2}
\begin{split}
    & P(Y=\hat{Y}|X_i=1, \mathbf{X}_{\mathcal{N}_i}=\mathbf{x}_{\mathcal{N}_i}, \mathbf{X}_{\mathcal{M}_i}=\mathbf{x}_{\mathcal{M}_i}) \\
    & \qquad = P(Y=\hat{Y}|X_i=1, \mathbf{X}_{\mathcal{N}_i}=\mathbf{x}_{\mathcal{N}_i}).
\end{split}
\end{equation}
Substituting this back into Equation~(\ref{eq1}):
\begin{equation}
\begin{aligned}
& P(Y=\hat{Y}|\mathrm{do}(X_i=1)) \\
& = \sum_{\mathbf{x}_{\mathcal{N}_i}} \sum_{\mathbf{x}_{\mathcal{M}_i}} P(Y=\hat{Y}|X_i=1, \mathbf{X}_{\mathcal{N}_i}=\mathbf{x}_{\mathcal{N}_i}) P(\mathbf{X}_{\mathcal{N}_i}=\mathbf{x}_{\mathcal{N}_i}, \mathbf{X}_{\mathcal{M}_i}=\mathbf{x}_{\mathcal{M}_i}) \\
& = \sum_{\mathbf{x}_{\mathcal{N}_i}} P(Y=\hat{Y}|X_i=1, \mathbf{X}_{\mathcal{N}_i}=\mathbf{x}_{\mathcal{N}_i}) \sum_{\mathbf{x}_{\mathcal{M}_i}} P(\mathbf{X}_{\mathcal{N}_i}=\mathbf{x}_{\mathcal{N}_i}, \mathbf{X}_{\mathcal{M}_i}=\mathbf{x}_{\mathcal{M}_i}).
\end{aligned}
\end{equation}
The inner summation over all configurations of $\mathbf{X}_{\mathcal{M}_i}$ is simply the marginal probability $P(\mathbf{X}_{\mathcal{N}_i}=\mathbf{x}_{\mathcal{N}_i})$.
\begin{equation}
\begin{aligned}
P(Y=\hat{Y}|\mathrm{do}(X_i=1)) = \sum_{\mathbf{x}_{\mathcal{N}_i}} &P(Y=\hat{Y}|X_i=1, \mathbf{X}_{\mathcal{N}_i}=\mathbf{x}_{\mathcal{N}_i}) \\
&\times P(\mathbf{X}_{\mathcal{N}_i}=\mathbf{x}_{\mathcal{N}_i}).
\end{aligned}
\end{equation}
This completes the proof.
\end{proof}

The remaining task is to compute the observational probabilities $P(Y=\hat{Y}|X_i=1, \mathbf{X}_{\mathcal{N}_i}=\mathbf{x}_{\mathcal{N}_i})$ and $P(\mathbf{X}_{\mathcal{N}_i}=\mathbf{x}_{\mathcal{N}_i})$. Inspired by \citet{bynum2024language}, we leverage LLMs to estimate these probabilities. Note that since each $X_j \in \mathbf{X}_{\mathcal{N}_i}$ is a Bernoulli variable, the summation is over all $2^{|\mathcal{N}_i|}$ permutations of their values. We construct prompts similar to the approach of \citet{halawi2024approaching} to query the LLM for these conditional and joint probabilities. The specific prompts used are detailed in Appendix~\ref{app: prompt} (e-f). For our experiments, we use a dependency window of $w=30$ days.

Importantly, trained on vast amounts of observational data, LLM excels at providing observational probabilities. However, LLMs are not strong enough to directly compute intervened probabilities~\cite{bynum2024language}. Our derivation provides the necessary bridge. By computing both terms and substituting them into Equation~\ref{eq:ace}, we can now calculate the \ace~scores for each news article.

\begin{figure*}
    \centering
    \includegraphics[width=2\columnwidth]{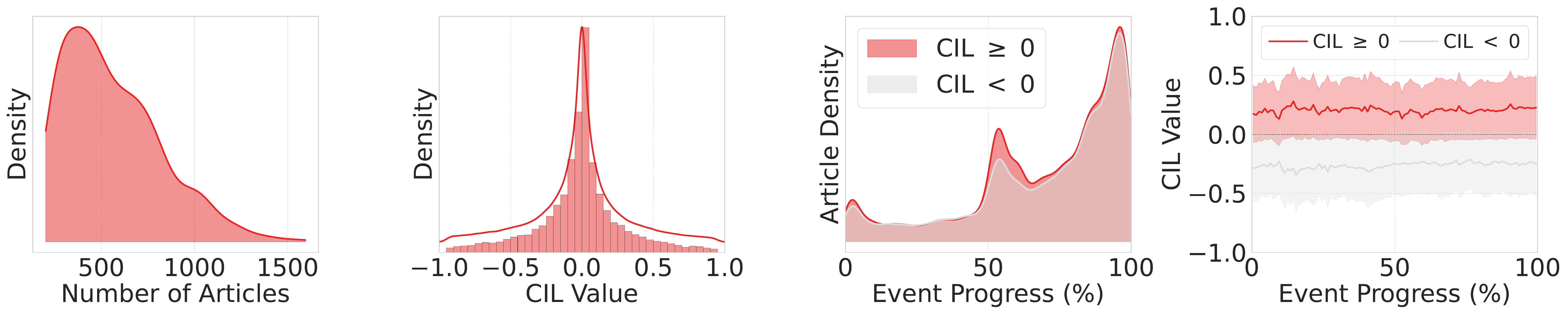}
    \caption{In-depth statistics on \ep. (a) Distribution of article counts per question. (b) Overall distribution of CIL values across all articles. (c) Article density along event progress (date ranging from the earliest to the latest news) for positive (CIL $\ge$ 0) and negative (CIL < 0). (d) Mean CIL values and standard deviations as a function of forecasting progress.}
    \label{fig:density}
\end{figure*}

\subsection{News Events Composition}
\label{sec:composition}

In practical computation of \ace~ (Eq. (\ref{eq:proposition})), we observe that the number of news articles within the $w$-day dependency window (we use $w=30$ days) can be extremely large.  
Directly enumerating all $2^{|\mathcal{N}_i|}$ value combinations of $\mathbf{X}_{\mathcal{N}_i}$ would lead to exponential computational costs, making the naive implementation infeasible.  
To address this issue, we design a \emph{News Events Composition} procedure to aggregate and compress related news items into fewer representative nodes before enumeration.  
This step preserves the key causal information needed for CIL computation while significantly reducing complexity.

The motivation behind this approach is that many news articles within a short period may describe similar or redundant events, differing only in details or reporting styles, and can be generalized to larger-granularity events.
Thus, we aggregate temporally and semantically similar articles into concise, representative summaries.

Formally, when computing \ace~for $X_i$, we proceed as follows:

\begin{itemize}[left=0.2cm]\setlength{\itemsep}{0pt}\setlength{\parskip}{0pt}\setlength{\topsep}{-5pt}
    \item \textbf{Individual summarization.}  
    We first employ the LLM to produce a concise, factual summary for each news article $X_j$ in the dataset.  
    These summaries capture the essential event semantics while discarding irrelevant text details, serving as a standardized basis for grouping.
    
    \item \textbf{Grouping within the dependency window.}  
    We identify the set $\mathcal{N}_i = \{ j \,|\, 0 \le T_i - T_j \le w \}$, i.e., all news events occurring within $w$ days before $X_i$.  
    We then divide these preceding events into groups of at most $10$ articles each, ordered chronologically.

    \item \textbf{Group-level relevance extraction.}  
    For each group, we concatenate the summaries of its articles into a single text block.  
    We then query the LLM to extract and condense only the information relevant to $X_i$ from this block.  
    This step removes unrelated details and focuses on causally pertinent content.

    \item \textbf{Clustering of extracted summaries.}  
    The relevance-extracted summaries from all groups are embedded into a semantic vector space (e.g., via sentence embeddings\footnote{https://sbert.net/}) and clustered based on semantic similarity.  
    Clustering captures redundancy among different groups, as reports on similar events may appear in multiple sources or days.

    \item \textbf{Final cluster summarization.}  
    For each cluster, we again use the LLM to summarize the shared content into a final, compact representation of that event cluster.  
    These final summaries constitute the reduced set of nodes that we enumerate when computing the \ace~ for $X_i$.
\end{itemize}

Through this multi-stage compression pipeline, we replace a large, potentially redundant set $\mathcal{N}_i$ with a much smaller set of semantically distinct representative events.  
This significantly decreases the number of permutations over $\mathbf{X}_{\mathcal{N}_i}$ in Equation~\eqref{eq:proposition}, thereby reducing the overall computational load, while still retaining the necessary causal context for accurate interventional probability estimation.

\subsection{Construction}
\label{sec: construction}

After calculating the \ace~for all pieces of news, we construct the benchmark with them. For each question, if it contains news articles with \ace > 0.7, we add the question to the chosen set \textit{L1}, otherwise to \textit{L2}.
We consider \textit{L1} to be the main part of our benchmark because answering the questions can be sufficiently supported by \textit{L1}. It can serve as an RAG benchmark.
While \textit{L2} lacks sufficient support to answer the questions, it also provides valuable information for prediction questions, but needs to be supplemented with additional information beyond the news. We use QwQ-32B~\cite{qwen2025qwen25technicalreport} to compute all observation probabilities and Qwen2.5-32B~\cite{qwen2025qwen25technicalreport} to do other data processing. Note that:

\noindent\textbf{Causality Assumptions.}
Our assumptions are rooted in general commonsense and aim to capture the dominant patterns in news-event relationships. We don't attempt to model global causality; instead, it suffices to model the causality required for the task with appropriate parameters.

\noindent\textbf{Probability Computing.} 
In pilot experiments, different LLMs provided slightly different scores when computing probabilities in \ace. Thus, we use a single LLM multiple times for reliable estimation. Later experiments showed that \ace~is model-agnostic: different models reach the same conclusions, validating this estimation method.

\subsection{Statistics and Properties of \ep}
\label{sec:stats}

We perform basic statistics of \ep, and, assisted by \ace, we further explore the key properties of the future forecasting task. Currently, we harvest 612 data points in \texttt{L1} and 43 data points in \texttt{L2}. The statistics of the crawled news articles are shown in Table~\ref{tab:stats}. During the construction process, we only discard obviously irrelevant news, meaning that we did not significantly alter the data distribution of valid news. Thus, the retained articles can reflect the real distribution of situations surrounding queried events.

For each question, we retain 100 top relevant news articles, resulting in an average of 559.51 articles per question. The average token count per article is 1249.88, which poses a challenge for methods that attempt to simply concatenate all news into a single prompt. The average \ace~score is $-0.013$, indicating that slightly more negative-supportive news exists compared to positive ones.

Beyond these summary statistics, we conduct more in-depth analyses, visualized in Figure~\ref{fig:density} and reveal several key insights:

\begin{itemize}[left=0.2cm]\setlength{\itemsep}{0pt}\setlength{\parskip}{0pt}\setlength{\topsep}{-5pt}
    \item \textbf{Article count distribution.} Figure~\ref{fig:density}(a) shows that most questions have between 400 and 800 associated articles, with a long tail of dense coverage for certain high-profile events. This indicates an uneven information supply across questions.
    
    \item \textbf{CIL value distribution.} As shown in Figure~\ref{fig:density}(b), the distribution of \ace~ values is sharply peaked around zero, indicating that most of the events are non-supportive. It's hard for methods to mine useful information for precise forecasting.
    
    \item \textbf{Temporal density of positive/negative articles.} Figure~\ref{fig:density}(c) illustrates article density over the course of event progress (date ranging from the earliest to the latest news) for positive (\ace~$\geq 0$) and negative (\ace~$< 0$) articles. We observe peaks at early stages, implying heightened reporting activity both in the initial phase and just before resolution.
    
    \item \textbf{Temporal evolution of sentiment.} As shown in Figure~\ref{fig:density}(d), the mean \ace~values remain relatively stable over time for both positive and negative groups, with only small oscillations. 
\end{itemize}

Overall, these statistical properties indicate that \ep~offers both rich contextual diversity and robust inferability, making it a challenging benchmark for forecasting systems that must reason over temporally distributed Web information.

\section{Experiments}

We first conduct experiments to show the validity of \ace~estimation and our benchmark in Section~\ref{sec:validity}. Then we evaluate various forecast methods on \ep~benchmark. We evaluate baselines of both Naive RAG and Agentic RAG.

\noindent\textbf{Naive RAG}
Naive RAG refers to the straightforward form of retrieval-augmented generation, 
where the system directly retrieves relevant documents from the knowledge base $\sX$ 
and conditions the generation process on them without iterative reasoning or tool use.
Formally, Naive RAG can be represented as:
\begin{equation}
\begin{aligned}
    \sR = \mathrm{Retrieve}(Q, B, \sX), Y = \mathrm{Generate}(Q, B, \sR),
\end{aligned}
\end{equation}
where $\sR$ denotes the set of retrieved articles, $\mathrm{Retrieve}$ is typically a dense or sparse retriever, and $\mathrm{Generate}$ is an LLM that outputs the probability estimation.
In this setting, there is no explicit multi-step reasoning over retrieval results; the model must encode 
all reasoning based solely on $\sR$. Prompt is in the Appendix prompt (g).

\noindent\textbf{Agentic RAG}
Agentic RAG extends Naive RAG by introducing an \emph{agent} that can iteratively retrieve, reason, and act on intermediate conclusions, 
often following the ReAct paradigm~\cite{yao2023react}. In ReAct, the model alternates between reasoning steps $\tau$, action steps $\alpha$ (tool invocation for retrieval), and receiving observation $o$ of the action, enabling multi-step evidence gathering before producing its final probability estimation: 
\begin{equation}
\begin{aligned}
(Q, \tau_1, \alpha_1, o_1, \tau_2, \alpha_2, o_2, \ldots, \tau_{L}, \alpha_{L}, o_{L})
\end{aligned}
\end{equation}
All agents adopt the ReAct framework.
Each agent is equipped with the same retrieval tool, prompts, and interaction structure, with the only variation being the underlying LLM.
The retrieval tool implements document retrieval: the agent issues a natural language query, which is converted into a dense vector representation via a fixed embedding model, followed by cosine similarity search over the news corpus $\sX$ to obtain the most relevant articles. Prompt is in the Appendix prompt (h). The tool is in the Appendix~\ref{sec:agentic_rag_tool}.

We show the evaluation results in Section~\ref{sec:naive_rag_analysis} and \ref{sec:agentic_rag}. Then we soly evaluate the reasoning abilities of each model in Section~\ref{sec:reasoning} and analyze the reasoning performances along the forecasting timeline in Section~\ref{sec:temporal}. We finally demonstrate cases in Section~\ref{sec:cases}.

% We first conduct experiments to show the validity of \ace~estimation and our benchmark in Section~\ref{sec:validity}. Then we evaluate the current retrieval and reasoning baselines on \ep~in Section~\ref{sec:performance}. Lastly, assisted by \ace, we conduct a temporal analysis on \ep~to provide insights into future forecasting systems in Section~\ref{sec: temporal}. We use \textbf{the cases to show the effectiveness of \ace~}in the Appendix~\ref{app: case}.

% \vspace{-2mm}
% \subsection{Evaluated Settings}
% \vspace{-2mm}

% For retrieval methods, we evaluate \texttt{Naive RAG}, \texttt{APP}~\cite{halawi2024approaching}, \texttt{Rankllama}~\cite{ma2024fine}, \texttt{HyDE}~\cite{gao-etal-2023-precise}.
% For reasoning methods, we include \texttt{ScrathPAD}~\cite{halawi2024approaching}, \texttt{CoT}~\cite{wei2022chain}, \texttt{Long-CoT}~\cite{openai2024o1}. Details are in the Appendix~\ref{app: baselines}. Since the news would be long, we pre-summarize each news and all methods use the same summarization in RAG.

\subsection{Validity of \ace~and \ep}
\label{sec:validity}

To validate the effectiveness of our proposed \ace~metrics, we design experiments to test their practical impact. 
The underlying intuition is that, if \ace~is indeed effective, forecasts based on the articles with the highest \ace~scores should significantly outperform forecasts obtained by directly answering the question without reference to these articles. 
To examine this, we conduct hypothesis tests to determine whether there is a statistically significant difference between the two approaches.

% 实验设置

%\section*{Experimental Setup}
We partition the entire news corpus into two disjoint subsets according to the distribution of CIL scores.
For each subset we evaluate two strategies:
(1) \textbf{Top CIL}: retrieve the top-20 articles ranked by CIL score and feed them into the generator;
(2)\textbf{without RAG}: let the generator condition solely on the query.\\
Every question is repeated 10 times to obtain stable estimates.

%\section*{Statistical Testing}
Let $X=(x_i)_{1\times 10n}$, where each $x_i$ is the Berrir score of a single response to one question under the \emph{with-RAG} strategy; analogously, let $Y=(y_i)_{1\times 10n}$ denote the scores produced \emph{without-RAG}. 
since every question is repeated 10 times, the vector contains $10n$ such individual scores.
Define the paired differences $Z=(x_i-y_i)_{1\times 10n}$.
We conduct one-sample $t$-tests on $Z$ with significance level $\alpha=0.05$ under two one-sided null hypotheses:
\[
\mathcal{H}_{0}^{\le}\colon\mu(Z)\le 0.077
\qquad\text{and}\qquad
\mathcal{H}_{0}^{\ge}\colon\mu(Z)\ge 0.027.
\]
Rejecting $\mathcal{H}_{0}^{\le}$ implies that the average Berrir score of \emph{with-RAG} significantly exceeds that of \emph{without-RAG} by at least $0.077$, i.e.\ the retrieved articles are beneficial for answering the question.
Rejecting $\mathcal{H}_{0}^{\ge}$ implies that the improvement is \emph{no more than} $0.027$, i.e.\ the articles provide negligible gain.

%\section*{Empirical Results}
\begin{table}[ht]
\centering
\label{tab:hypo}
\begin{tabular}{lcccc}
\toprule
Subset & Hypothesis & Decision & $p$-value & $\bar{d}$ \\
\midrule
L1 & $\mathcal{H}_{0}^{\le}\colon \mu(Z)\le 0.077$ & Rejected & 0.0264 & \multirow{2}{*}{0.0835} \\
   & $\mathcal{H}_{0}^{\ge}\colon \mu(Z)\ge 0.027$ & Not rejected & 1.0000 & \\[4pt]
L2 & $\mathcal{H}_{0}^{\le}\colon \mu(Z)\le 0.077$ & Not rejected & 1.0000 & \multirow{2}{*}{0.0176} \\
   & $\mathcal{H}_{0}^{\ge}\colon \mu(Z)\ge 0.027$ & Rejected & 0.0474 & \\
\bottomrule
\end{tabular}
\caption{Hypothesis-testing results for subsets L1 and L2. $\bar{d}=\bar{x}-\bar{y}$ denotes the mean improvement of \emph{Top CIL} over \emph{without RAG}.}
\end{table}
%\vspace{-4mm}
% 对结果的分析的结论

The statistical test results reveal that, under the \texttt{L1} setting, 
the null hypothesis $\mathcal{H}_0^{\le}$ is rejected ($p = 0.0264$), indicating that forecasts constructed from the top-$20$ \ace~scored articles achieve a significantly larger improvement (over~$7.7$) compared with directly answering the question.  In contrast, under the L2 setting, $\mathcal{H}_0^{\le}$ cannot be rejected ($p = 1.0000$), 
suggesting that the improvement over direct answers is not statistically significant.  
This pattern demonstrates that \ace~can successfully identify highly supportive news articles. 
And the construction of \ep~is valid based on this metric.

\subsection{Reasoning Performances}
\label{sec:reasoning}

\begin{figure}
    \centering
    \includegraphics[width=1\columnwidth]{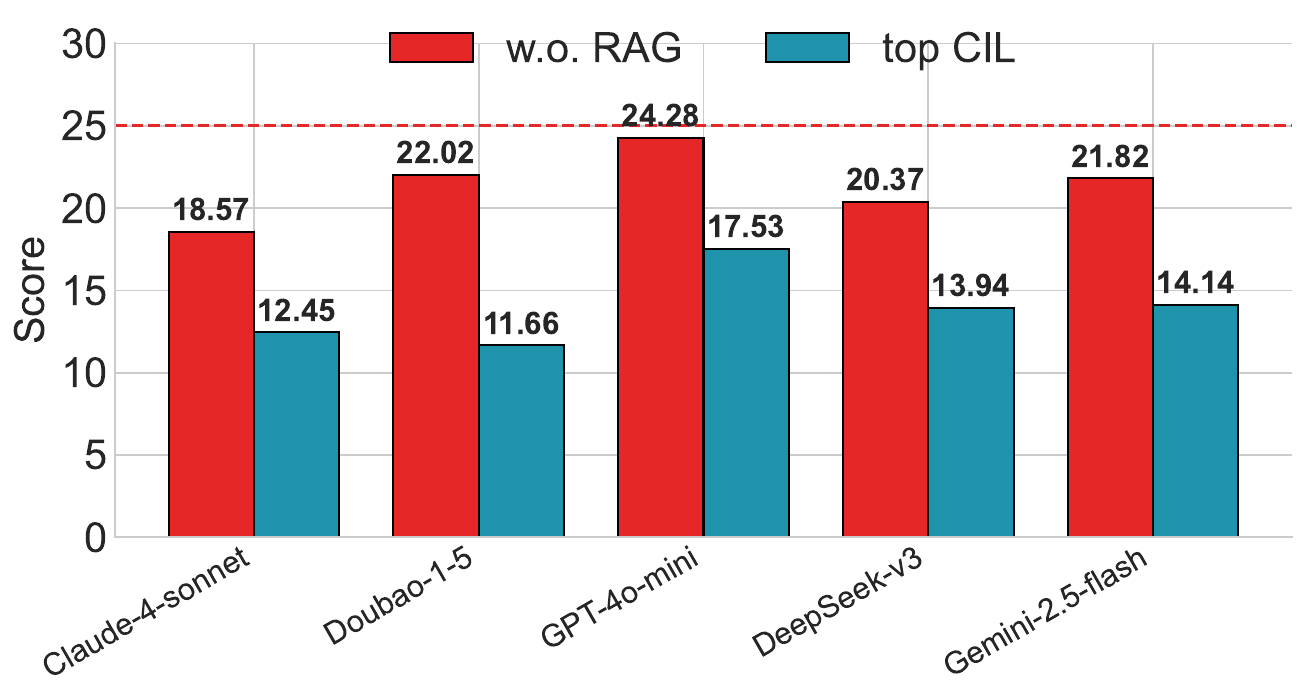}
    \caption{Reasoning ability evaluation. The red line stands for random results which is 25.0.}
    \label{fig:reasoning}
\end{figure}

We select the top-10 \ace~articles of each question for prediction, and compare to performances without RAG. The results are shown in Figure~\ref{fig:reasoning}.
From the experimental results, it is clear that the Brier Scores of the top-10 CIL selection are significantly better than those achieved without RAG for all tested models. This further demonstrates the effectiveness of the \ace~metric in identifying high-quality articles that are capable of boosting forecasting performance. 

Moreover, the strong performance observed in the top-10 CIL setting suggests that the degree of answer leakage within the dataset is minimal. If substantial leakage were present, RAG-assisted prediction would not exhibit such improvements over the baseline, as the retrieved content would simply repeat the ground-truth answers rather than genuinely aiding reasoning. Therefore, these results provide additional evidence that our dataset maintains integrity while allowing meaningful performance gains through retrieval.

\subsection{Evaluation On Naive RAG Baselines}
\label{sec:naive_rag_analysis}

We evaluate a set of naive Retrieval-Augmented Generation (RAG) baselines over the constructed \ep{} dataset to establish a performance reference and to examine the practical challenges of the task.  
For the generator component, we select several representative Large Language Models: Claude-sonnet-4, Doubao-1.5, GPT-4o-mini, DeepSeek-v3, and Gemini-2.5-flash.  
The retrieval component employs seven popular open-source embedding models: all-MiniLM-L6-v2 (AM), msmarco-distilbert-cos-v5 (MDC), msmarco-MiniLM-L6-cos-v5 (MM), msmarco-distilbert-dot-v5 (MDD), msmarco-bert-base-dot-v5 (MBD), instructor-large (IL), and instructor-base (IB). 
We evaluate each LLM–embedding pair with retrieval sizes $n=10$ and $n=20$, reporting Brier Scores (lower is better). 
The results are shown in Table~\ref{tab:naive_rag_results}.
From the results, we make the following takeaways:

\noindent\textbf{Limited capability of Naive RAG.}  
Across all evaluated configurations, introducing naive RAG does not consistently outperform the ``w.o. RAG'' (no retrieval) baseline, and in many cases even leads to performance degradation.  
This indicates that simply appending retrieved documents to the LLM input, without any filtering, temporal alignment, or causal reasoning, is insufficient for the \ep{} forecasting task.  
The retrieved context often contains redundant or irrelevant information, and may conflict with the model's internal knowledge, which can confuse the generator and inflate the Brier Score.  
Moreover, the challenges of identifying truly predictive evidence from historical data suggest that naive RAG lacks mechanisms to reason about event timelines, domain-specific causal links, or uncertainty, all of which are essential for accurate forecasting.

\noindent\textbf{Small differences across current embedding models.}
Changing among the seven embedding models yields only modest performance variation for a given LLM.  
This suggests that current general-purpose retrievers are not well suited for temporally and causally grounded forecasting, and that more task-specific embedding models may be needed.
% \noindent\textbf{Small differences across embedding models.}
% Switching among the seven models causes only minor performance changes, suggesting that general-purpose retrievers are suboptimal for temporally or causally grounded forecasting, and task-specific embeddings may be required.

%\vspace{-1mm}
\noindent\textbf{Increasing retrieval size has limited benefit.}
Increasing from $n=10$ to $n=20$ retrieved documents seldom improves performance and sometimes even degrades it.  
Merely adding more documents can introduce noise and increase reasoning difficulty for the LLM.  
High-quality selection of truly relevant evidence is more important than raw retrieval quantity.

Overall, naive RAG configurations underperform on \ep{}, and the results highlight the necessity of causally aware retrieval strategies and reasoning methods tailored to forecasting.

\begin{table*}[t]
\centering
\setlength\tabcolsep{6pt}
\begin{tabular}{lcccccccc}
\hline
 & w.o. RAG & AM & MDC & MM & MDD & MBD & IL & IB \\
\hline
Claude-4-sonnet & 18.57& 19.00/19.49 & 19.15/18.72 & 18.12/18.00 & 19.15/19.64 & 18.41/18.87 & 18.24/19.45 & 19.69/19.14 \\
Doubao-1-5 & 22.02 & 20.73/21.01 & 21.45/21.00 & 22.72/21.09 & 20.53/20.25 & 21.89/21.23 & 20.33/21.18 & 21.85/22.05 \\
GPT-4o-mini & 24.28 & 27.17/27.20 & 28.32/28.92 & 28.34/27.89 & 29.41/28.95 & 29.08/30.34 & 27.52/28.06 & 29.10/28.46 \\
DeepSeek-v3 & 20.37 & 21.56/21.48 & 21.94/21.74 & 22.22/21.04 & 21.33/22.60 & 22.10/22.27 & 21.60/22.96 & 23.08/22.55 \\
Gemini-2.5-flash & 21.82 & 21.63/21.48 & 23.69/22.19 & 22.58/23.63 & 21.11/22.68 & 22.06/20.15 & 21.54/21.16 & 21.99/22.06 \\
\hline
\end{tabular}
\caption{Naive RAG on the \ep{} dataset. Lower is better. Each cell reports the score with $n=10$/$n=20$ articles. }
\label{tab:naive_rag_results}
\end{table*}

\subsection{Agentic RAG Performances}
\label{sec:agentic_rag}

\begin{table*}
\centering
\setlength\tabcolsep{10pt}
\begin{tabular}{lcccccccc}
\hline
& w.o. RAG & \multicolumn{3}{c}{AM} & \multicolumn{3}{c}{MDC} \\
\cmidrule(lr){3-5} \cmidrule(lr){6-8}
& BS & BS ($\delta$) & Step & Thought & BS ($\delta$) & Step & Thought\\
\hline
\hline
Claude-4-sonnet & 18.57 & 18.65 {\color[HTML]{E62727}(+0.08)} & 3.67 & 195.65 & 17.89 {\color[HTML]{1E93AB}(-0.68)} & 3.58 & 211.99 \\
Claude-3.5-sonnet & 22.22 & 21.53 {\color[HTML]{1E93AB}(-0.69)} & 1.05 & 169.75 & 21.54 {\color[HTML]{1E93AB}(-0.68)} & 1.02 & 202.83 \\
Doubao-1.5 & 20.20 & 20.40 {\color[HTML]{E62727}(+0.20)} & 1.02 & 129.97 & 24.53 {\color[HTML]{E62727}(+4.33)} & 1.08 & 166.08 \\
Gemini-25-pr0 & 21.41 & 20.51 {\color[HTML]{1E93AB}(-0.90)} & 0.95 & 139.64 & 19.26 {\color[HTML]{1E93AB}(-2.15)} & 0.95 & 151.01 \\
GPT-4.1 & 21.64 & 21.14 {\color[HTML]{1E93AB}(-0.50)} & 0.61 & 91.24 & 18.44 {\color[HTML]{1E93AB}(-3.20)} & 0.18 & 101.94 \\
GPT-4o & 22.31 & 23.33 {\color[HTML]{E62727}(+1.02)} & 1.02 & 102.64 & 22.47 {\color[HTML]{E62727}(+0.16)} & 1.04 & 125.38 \\
% Grok-4 & 22.31 & 18.82 {\color[HTML]{1E93AB}(-3.49)} & 4.32 & 170.29 & 18.18 {\color[HTML]{1E93AB}(-4.13)} & 4.35 & 184.79 \\
\hline
\end{tabular}
\caption{Performances on Agentic RAG. Lower is better. $\delta$ stands for differences between Agentic RAG and w.o. RAG. Thought is the average thought length of each tool call.}
\label{tab:agentic_rag_comparison}
\end{table*}

We also conduct experiments on Agentic RAG methods. AM and MDC are different embedding models used in the tool.
From Table~\ref{tab:agentic_rag_comparison}, Agentic RAG achieves consistent gains over the w.o. RAG baseline and outperforms naive RAG (Section~\ref{sec:naive_rag_analysis}) for several LLMs, with models like GPT-4.1, and Gemini-2.5-pro showing Brier Score reductions exceeding $-2.0$. 
Unlike naive RAG’s simple document concatenation, Agentic RAG allows the LLM to iteratively retrieve, inspect, and integrate evidence, improving temporal and causal relevance while reducing noise.  
Performance gains are often linked to deeper reasoning, as indicated by higher \textit{Step} counts and longer \textit{Thought} lengths.  
These results highlight Agentic RAG as a promising direction for forecasting, combining adaptive retrieval with structured reasoning to better exploit external knowledge.

\subsection{Temporal Analysis}
\label{sec:temporal}

\begin{figure}
    \centering
    \includegraphics[width=1\columnwidth]{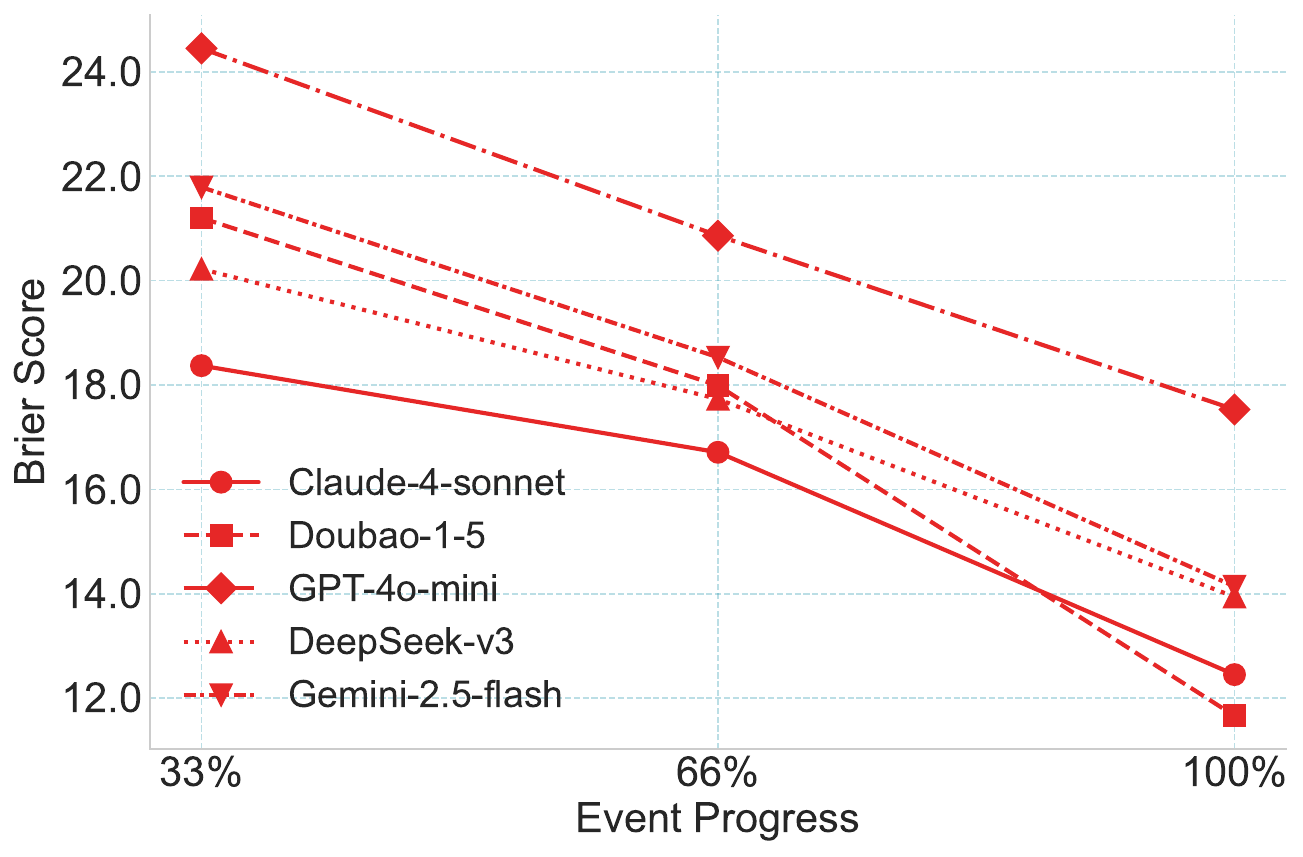}
    \caption{Temporal analysis. Results are on top-10 \ace~articles on various forecasting progress.}
    \label{fig:temporal}
\end{figure}

Future forecasting is a continuous process that begins when the question is posed and ends when the question is answered. The earlier the answer can be predicted, the more valuable it is. We investigate the system's forecasting at different times. Similar to Section~\ref{sec:stats}, we compute the progress in the whole forecasting. We represent the progress of each news by the percentage of its date in the forecasting. 
We run different models based on the top-10 \ace~articles in various prediction progress.
The results are in Figure~\ref{fig:temporal}.
As the prediction process progresses, the difficulty of prediction decreases, but early predictions still face great difficulties.

% \textcircled{1} We find significant potential in the early-time future forecasting. The \texttt{CIL-High} at 20\% progress performs even better than \texttt{Naive RAG} at 100\%. It indicates that if we have a sufficiently powerful retrieval method, we can expect to achieve effective predictions at the early stages of event development.
% This finding applies to both scenarios where evidence is sufficient and where it is insufficient.

% \textcircled{2} When the forecasting progress precedes, there would be news that is harmful for prediction. We find that during the progress of forecasting, the performances of some methods fluctuate. 
% And the \ace~ of the \texttt{Naive RAG} stops increasing at 60\%. This is consistent with the conclusions in Section~\ref{sec:stats}. 
% It shows a desired prediction system should be aware of negative evidence and can self-correct in the retrieval and reasoning process.

%\vspace{-3mm}
\section{Related Work}
\label{sec:related_work}

\subsection{Future Forecasting and Benchmarks}

Previous research on future forecasting benchmarks has evolved in different paradigms, each addressing different aspects of the task. Early benchmarks, such as MCNC~\cite{granroth2016happens}, SCT~\cite{mostafazadeh2017lsdsem}, and CoScript~\cite{yuan2023distilling}, focused on script learning and common sense reasoning in synthetic scenarios. Although these data sets facilitated structured reasoning, they lacked real-world applicability and grounding in factual news. Time series datasets such as GDELT~\cite{leetaru2013gdelt} and ICEWS~\cite{schrodt2012icews} introduced real-world event tracking but did not formalize prediction as a retrieval-augmented reasoning task or ensure answerability. Later works, such as ECARE~\cite{du2022care}  and EV2~\cite{tao2024comprehensive}, 
%advanced event reasoning but remained confined to settings without real-world grounding. 
advanced event reasoning has made significant progress in understanding abstract or synthetic scenarios but remains largely confined to settings without real-world grounding, limiting its applicability to practical forecasting or causal inference tasks.

With the rise of LLMs, recent benchmarks such as \citet{halawi2024approaching}, OpenEP~\cite{guan2024openep}, and ForecastBench~\cite{karger2024forecastbench} shifted the focus to real-world questions and news-based search. However, these datasets suffer from two critical limitations: (1) they lack explicit validation of inferability, allowing questions with insufficient supporting evidence to persist, and (2) they prioritize dynamic data sources over reproducibility, risking inconsistent evaluations due to evolving news archives.  
\ep~ addresses these gaps by filtering via the introduced Causal Intervened Likelihood estimation. 
% We show the benchmark comparison in Table~\ref{tab:comparison}.

% Before the era of LLMs, future forecasting benchmarks evaluate more of the commonsense reasoning ability of models. These benchmarks are usually constructed by human annotators which naturally ensure the predictive of them~\cite{granroth2016happens, mostafazadeh2017lsdsem, yuan2023distilling, tao2024comprehensive}. However, limited to simplified scenarios, these benchmarks don't evaluate real-world future forecastings. 

% In the era of LLMs, researchers develop real-world future forecasting behcmarks~\cite{halawi2024approaching, guan2024openep, karger2024forecastbench}. They formulate the prediction task as an RAG with hidden rationales which are more challenging than those before. However, the merit of real-world prediction questions incurs a problem of predictive validation. The benchmarks would be non-predictable without it. 
% Moreover, if the grounded news articles are not fixed, it may lead to a situation where researchers cannot download the same news articles when using the dataset, and thus the results cannot be reproductive.
% Our \ep~is the first one consisting of real-world questions with predictive validation. We show the comparison in Table~\ref{tab:comparison}.

%\vspace{6mm}

\subsection{RAG and Benchmarks}
\noindent\textbf{Foundational QA Datasets for RAG}: Traditional QA datasets, including MMLU~\cite{hendrycks2021mmlu}, StrategyQA~\cite{geva2021strategyqa}, ASQA~\cite{stelmakh2022asqa}, Multi-HopQA~\cite{ho20202wikimultihopqa}, and 2WikiMultiHopQA~\cite{ho20202wikimultihopqa}, are adapted to evaluate RAG systems. These datasets, grounded in knowledge bases like Wikipedia, form the basis for RAG evaluation.

\noindent\textbf{Domain-Agnostic}: RAGBench~\cite{friel2024ragbench} is a multi-domain benchmark across biomedical, legal, customer support, and finance domains. CRAG~\cite{crag} provides a factual QA benchmark across five domains, simulating web and knowledge graph search.

\noindent\textbf{Domain-Specific}: Domain-specific benchmarks include LegalBench-RAG~\cite{legalbenchrag}, WeQA~\cite{meyur2024weqa}, PubHealth~\cite{zhang2023pubhealth}, and MTRAG~\cite{mtag}. These benchmarks address niche applications and improve evaluation precision in domains.

\noindent\textbf{Capability-Oriented}: RGB~\cite{benchmarking_large_language_models_rag} evaluates four RAG capabilities: noise robustness, negative rejection, information integration, and counterfactual robustness. TRIAD~\cite{zong2024triadframeworkleveragingmultirole} assesses retrieval quality, fidelity, and utility through a three-dimensional framework.

In this work, we focus on the inferability of RAG benchmarks, a key property for domain-specific and real-world scenarios. Our method can be generalized to other domains.

\section{Conclusion}  
We address the challenge of building the inferable RAG benchmark for evaluating future forecasting systems by introducing \ep. It is rigorously validated for inferability by our Causal Intervened Likelihood (\ace) estimation. By leveraging causal inference to quantify the inferability of prediction questions based on their associated news articles, \ep~ensures that questions are answerable through retrieved rationales, thereby providing a more accurate assessment of the model capabilities. Experimental validation confirms the effectiveness of \ace~in correlating with system performance, while evaluations of state-of-the-art systems on \ep~reveal key strengths and limitations, particularly in retrieval and reasoning. This work establishes a basis for the development of more nuanced models. With ongoing updating, \ep~ensures the inferable evaluation in driving progress towards AI-powered forecasting.

% \section*{Limitations}
% In this work, we evaluate methods of retrieval and reasoning disentangling. However, entangled methods could further improve future forecasting. We leave it to future work.

% \section*{Ethics Statement}
% This dataset is strictly for non-commercial research purposes under the following conditions: 1) Restricted Application Scope: All narrative scenarios contained herein are intended solely for academic exploration of future forecasting methodologies. Any utilization for purposes involving defamation, harassment, malicious targeting, or other unethical practices is expressly prohibited.   
% 2) Prohibited Misinterpretation: Statistical patterns derived from this resource should not be interpreted as deterministic predictions of real-world events. 
% 3) Accountability Framework: The creators explicitly disclaim liability for consequences arising from dataset misuse, including but not limited to algorithmic bias propagation, privacy infringements, or sociotechnical harms caused by improper application.  

\bibliographystyle{ACM-Reference-Format}
\bibliography{main}

% \newpage
\appendix

% \newpage
\clearpage

\begin{figure*}[ht]
    \centering
    \includegraphics[width=2\columnwidth]{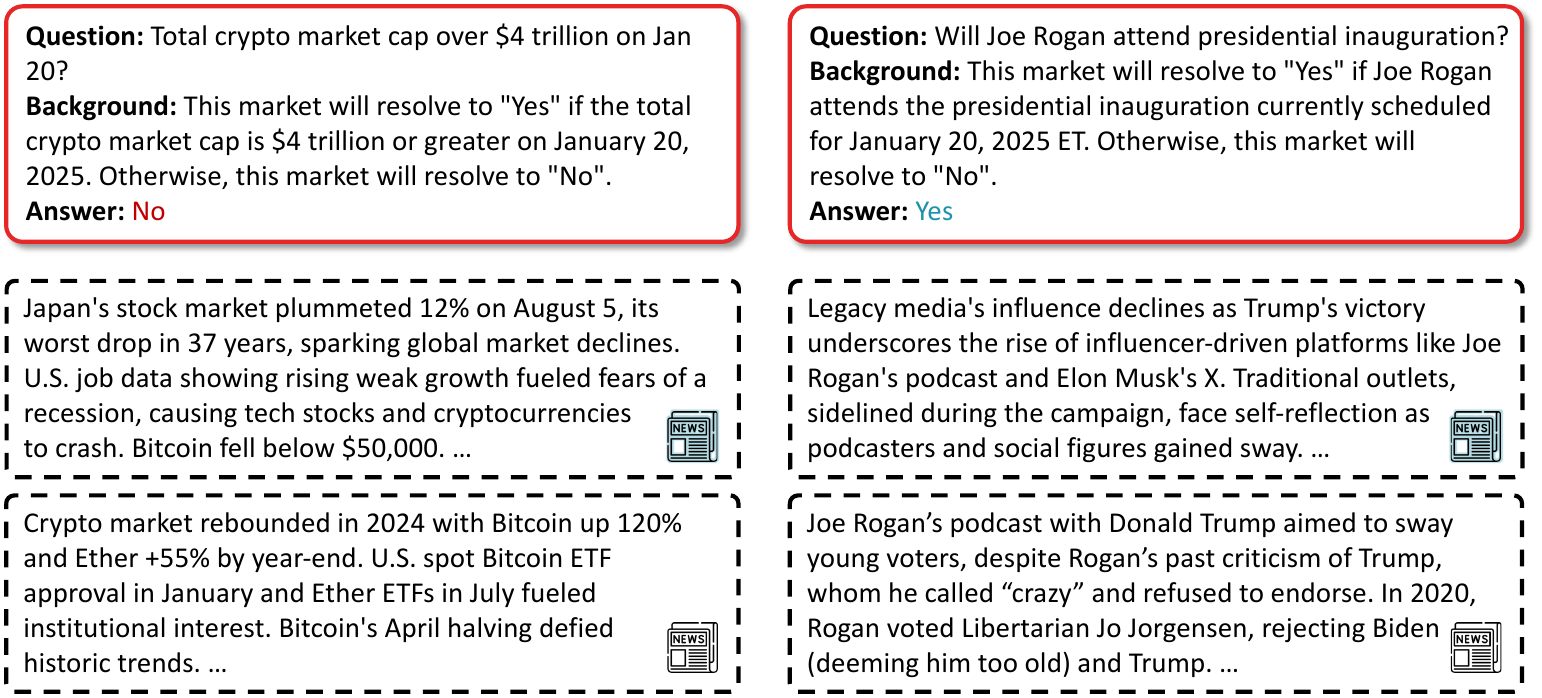}
    \caption{Case studies. The supportive and non-supportive articles are selected by \ace~scores.}
    \label{fig:cases}
\end{figure*}

\section*{Appendix}
\label{sec: appendix}

\section*{Case Studies}
\label{sec:cases}

To evaluate the proposed supportiveness metric in RAG, we consider two representative cases shown in Figure~\ref{fig:cases}. The first question asks whether the total crypto market cap will exceed \$4 trillion on January 20, 2025. Among the retrieved articles, the blue-highlighted news discusses substantial rebounds in Bitcoin and Ether alongside ETF approvals—information directly indicative of a “Yes” outcome. Conversely, the white-highlighted news describes global market declines and cryptocurrency crashes, contradicting the supportive stance. Our metric assigns higher scores to the supportive article, reflecting its stronger evidential link to the answer.

The second question concerns whether Joe Rogan will attend the presidential inauguration. The supportive article details Rogan’s past political activities and his collaboration with Donald Trump during the campaign, which strengthens the likelihood of his attendance. In contrast, the non-supportive article mentions Rogan only in the context of broader shifts in media influence, without providing any concrete statement or event suggesting his presence at the inauguration. Our metric correctly ranks the supportive article higher, aligning with the ground truth answer “Yes”.

Across both cases, the metric effectively distinguishes articles offering direct, answer-relevant evidence from those that are tangential or context-only, thereby enhancing RAG answer accuracy.

\section*{Construction Details}
During constructing, we use \texttt{gpt-4o-mini-2024-07-18} for all LLM callings. We set window size $w$ to 3 which is enough large in our pilot study. For computing each probability in \ace, we call twice \texttt{gpt-4o-mini-2024-07-18} and get the average score. The constructing prompts we use are shown in prompts (a-f).

\section*{Prompts} \label{app: prompt}
We list all prompts in the following Figures (a-h).

\newpage
\section*{Agentic RAG Tool}
\label{sec:agentic_rag_tool}

We show the article retrieval tool for all agentic RAG methods:

% 3. 在文档中使用此环境（在 \begin{document}...\end{document} 之间）
\begin{lstlisting}[style=CustomStyle, caption={Python Class Definition for Agentic RAG Tool}, label={lst:rag_tool_custom}]
class AgenticRAGTool:
    name = "query_rag"
    description = (
        "To search in a previously built RAG index to find the most relevant chunks of text."
    )
    parameters = [
        {
            'name': 'query',
            'type': 'string',
            'description': 'the query text to search for relevant text chunks.',
            'required': True
        },
        {
            'name': 'top_k',
            'type': 'integer',
            'description': 'the number of top relevant chunks to retrieve (default is 3).',
            'required': False
        }
    ]
\end{lstlisting}

\begin{figure*}
\centering
\begin{AcademicBox} [\texttt{(a) Entity Query Generation}]
    \small
    \vspace{1mm}

    I will provide you with a forecasting question and the background information for the question. Extract the named entities, events of the question. Each entity and event are up to 5 words. The named entities can only be people, organizations, countries, locations while can not be date or time. Put all result items in a list that I can parse by JSON as ["entity 1", "entity 2", "event 1", "event 2", ...].\\

    Question: $Q$\\
    
    Question Background: $B$ \\
    
    Question Date: $D$ \\
    
    Output:

\end{AcademicBox} 
\end{figure*}

\begin{figure*}
\centering
\begin{AcademicBox} [\texttt{(b) Resolving Steps Query Generation}]
    \small
    \vspace{1mm}

    I will provide you with a forecasting question and the background information for the question. I will then ask you to generate short search queries (up to {max words} words each) that I'll use to find articles on Google News to help answer the question. The articles should be mainly about event arguments such as subjects, objects, locations, organizations of the events in question and background information.
    You must generate this exact amount of queries: {num keywords}. Put all result items in a list that I can parse by JSON as ["step 1", "step 2", "step 3", ...].\\
    
    Question: $Q$\\
    
    Question Background: $B$ \\
    
    Question Date: $D$ \\
    
    Output:

\end{AcademicBox} 
\end{figure*}

\begin{figure*}
\centering
\begin{AcademicBox} [\texttt{(c) Similar Events Query Generation}]
    \small
    \vspace{1mm}

    I will provide you with a forecasting question and the background information for the question. I will then ask you to generate short search queries (up to {max words} words each) that I'll use to find articles of similar events on Google News to help answer the question. The similar events are events happened on other similar entities in the history. Or events happended on question entities but on other date.
    You must generate this exact amount of queries: {num keywords}. Put all result items in a list that I can parse by JSON as ["event 1", "event 2", "event 3", ...].\\
    
    Question: $Q$\\
    
    Question Background: $B$ \\
    
    Question Date: $D$ \\
    
    Output:

\end{AcademicBox} 
\end{figure*}

\begin{figure*}
\centering
\begin{AcademicBox} [\texttt{(d) News Article Relevance Rating}]
    \small
    \vspace{1mm}

    Please consider the following forecasting question and its background information.
    After that, I will give you a news article and ask you to rate its relevance with respect to the forecasting question. \\
    
    \#\#\# Question: $Q$\\
    
    \#\#\# Question Background: $B$\\
    
    \#\#\# Article: \{article\}\\
    
    Please rate the relevance of the article to the question, at the scale of 1-6\\
    1 -- completely irrelevant\\
    2 -- slightly relevant\\
    3 -- somewhat relevant\\
    4 -- relevant\\
    5 -- highly relevant\\
    6 -- completely relevant directly reveals the answer\\
    
    Guidelines:\\
    - scale 1-5 don't reveal the answer, only scale 6 reveals the answer.\\
    - If the article has events of similar types which may happened on different subjects, it also consider relevant to the question to some degree.\\
    - You don't need to access any external sources. Just consider the information provided.\\
    - If the text content is an error message about JavaScript, paywall, cookies or other technical issues, output a score of 1.\\
    
    Your response should look like the following, don't generate other words in the response:\\
    Rating: 

\end{AcademicBox} 
\end{figure*}

\vspace{-4mm}
\begin{figure*}
\centering
\begin{AcademicBox} [(e) Probability \ensuremath{P(Y=\hat{Y}|X_i=1,\mathbf{X}_{\mathcal{N}_i}=\mathbf{x}_{\mathcal{N}_i})}]
    \small
    \vspace{1mm}

    \#\#\# Given some event: \{events\}\\

    \#\#\# Most important event: \{current event\}\\

    \#\#\# Instructions: Based on the information, especially the most important event, compute the probability that the answer for the following question is yes. Question: $Q$\\
    
    \#\#\# The mentioned named entities are real in life, no need to check that. You compute the probability based on your world knowledge about the mentioned named entities and the internal logic in the situation. the Output your answer (a probability number between 0 and 1).

\end{AcademicBox} 
\end{figure*}

\vspace{-4mm}
\begin{figure*}
\centering
\begin{AcademicBox} [(f) Probability \ensuremath{P(\mathbf{X}_{\mathcal{N}_i}=\mathbf{x}_{\mathcal{N}_i})}]
    \small
    \vspace{1mm}

    \#\#\# Given a situation describing some events: \{events\} \\ 

    \#\#\# Instructions: Reason the probability of the situation. The mentioned named entities are real in life, no need to check that. You compute the probability based on your world knowledge about the mentioned named entities and the internal logic in the situation. Output your answer (a probability number between 0 and 1).

\end{AcademicBox} 
\end{figure*}

\vspace{-4mm}
\begin{figure*}
\centering
\begin{AcademicBox} [(g) Naive RAG]
    \small
    \vspace{1mm}
    
    Your task is to answer the prediction question. The question asks about whether a future event would happen or not. Output a probability between 0 and 1 representing the likelihood of the event happening. Output the probability in the following format:\\
    <Probability>...</Probability>\\
    
    Use the provided articles as context to inform your prediction. The articles are provided below.\\
    
    Question: $Q$ \\
    Background: $B$ \\
    Create date of the question: $D$ \\
    
    Context Articles: \{retrieved articles\} \\

\end{AcademicBox} 
\end{figure*}

\vspace{-10mm}
\begin{figure*}
\centering
\begin{AcademicBox} [(h) Agentic RAG]
    \small
    \vspace{1mm}
    
    You are a prophet agent who can answer a query based on the data and the background information. Use the tools below to help you answer the question. \\

    Your task is to answer the prediction question. The question asks about whether a future event would happen or not. Output a probability between 0 and 1 representing the likelihood of the event happening. Output the probability in the following format:\\
    <Probability>...</Probability>\\
    
    Question: $Q$ \\
    Background: $B$ \\
    Create date of the question: $D$ \\

\end{AcademicBox} 
\end{figure*}


\end{document}